\documentclass[sigconf]{acmart}

\AtBeginDocument{%
  \providecommand\BibTeX{{%
    \normalfont B\kern-0.5em{\scshape i\kern-0.25em b}\kern-0.8em\TeX}}}

\acmConference[KDD '21]{KDD '21: ACM SIGKDD International Conference on Knowledge Discovery \& Data Mining}{August 14--18, 2021}{Singapore}

\setcopyright{rightsretained}

\def\xb{{\mathbf x}}

\def\thetab{{\boldsymbol\theta}}

\def\omegab{{\boldsymbol\omega}}

\def\Real{{\mathbb{R}}}

\def\Gcal{\mathcal{G}}

\def\Bcal{\mathcal{B}}
\def\Hcal{\mathcal{H}}

\def\Ccal{\mathcal{C}}
\def\Ncal{\mathcal{N}}
\def\Kcal{\mathcal{K}}

\def\Acal{\mathcal{A}}
\def\Dcal{\mathcal{D}}

\def\Pcal{\mathcal{P}}

\def\Xcal{\mathcal{X}}

\def\Exp{{\mathbb{E}}}

\def\defin{\triangleq}

\begin{document}
\fancyhead{} 
\title{Amazon SageMaker Automatic Model Tuning: \\ Scalable Gradient-Free Optimization}

\author{Valerio Perrone$^1$, Huibin Shen, Aida Zolic, Iaroslav Shcherbatyi, Amr Ahmed \\ Tanya Bansal, Michele Donini, Fela Winkelmolen$^*$, Rodolphe Jenatton$^*$ \\ Jean Baptiste Faddoul, Barbara Pogorzelska, Miroslav Miladinovic \\ Krishnaram Kenthapadi, Matthias Seeger, C\'{e}dric Archambeau}

\affiliation{%
 \institution{Amazon Web Services}
 }

\thanks{$^1$Correspondence to: Valerio Perrone, <vperrone@amazon.com>.}  %
\thanks{$^*$Work done while at Amazon Web Services.}  %

\renewcommand{\shortauthors}{Perrone et al.}

\begin{abstract}
Tuning complex machine learning systems is challenging. Machine learning typically requires to set hyperparameters, be it regularization, architecture, or optimization parameters, whose tuning is critical to achieve good predictive performance. To democratize access to machine learning systems, it is essential to automate the tuning. This paper presents Amazon SageMaker Automatic Model Tuning (AMT), a fully managed system for gradient-free optimization at scale. AMT finds the best version of a trained machine learning model by repeatedly evaluating it with different hyperparameter configurations. It leverages either random search or Bayesian optimization to choose the hyperparameter values resulting in the best model, as measured by the metric chosen by the user. AMT can be used with built-in algorithms, custom algorithms, and Amazon SageMaker pre-built containers for machine learning frameworks. We discuss the core functionality, system architecture, our design principles, and lessons learned. We also describe more advanced features of AMT, such as automated early stopping and warm-starting, showing in experiments their benefits to users.
\end{abstract}


\copyrightyear{2021} 
\acmYear{2021} 
\acmConference[KDD '21]{Proceedings of the 27th ACM SIGKDD Conference on Knowledge Discovery and Data Mining}{August 14--18, 2021}{Virtual Event, Singapore}
\acmBooktitle{Proceedings of the 27th ACM SIGKDD Conference on Knowledge Discovery and Data Mining (KDD '21), August 14--18, 2021, Virtual Event, Singapore}\acmDOI{10.1145/3447548.3467098}
\acmISBN{978-1-4503-8332-5/21/08}


\maketitle

\section{Introduction}

In modern machine learning, complex statistical models with many free parameters are fit to data by way of automated and highly scalable algorithms. For example, weights of a deep neural network are learned by stochastic gradient descent (SGD), minimizing a loss function over the training data. Unfortunately, some remaining {\em hyperparameters (HPs)} cannot be adjusted this way, and their values can significantly affect the prediction quality of the final model. In a neural network, we need to choose the learning rate of the stochastic optimizer, %
regularization constants, the type of activation functions, and architecture parameters such as the number or the width of the different layers. In Bayesian models, priors need to be specified, %
while for random forests or gradient boosted decision trees, %
the number and maximum depth of trees are important HPs.

The problem of hyperparameter tuning can be formulated as the minimization of an objective function $f: \Xcal \rightarrow \Real$, where $\Xcal$ denotes the space of valid HP configurations, while the value $f(\xb)$ corresponds to the metric we wish to optimize. We assume that $\xb = [x_1, \dots, x_d]$, where $x_j \in \Xcal_j$ is one of the $d$ hyperparameters, such that $\Xcal = \Xcal_1\times\ldots\times \Xcal_d$. For example, given some $\xb \in \Xcal$, $f(\xb)$ may correspond to the held-out error rate of a machine learning model when trained and evaluated using the HPs $\xb$. In practice, hyperparameter optimization requires addressing a number of challenges. First, we do not know the analytical form of the function $f$ ({\it i.e.}, it can only be observed through evaluations) and is thus difficult to optimize as we cannot compute its gradients. Second, evaluations of $f(\xb)$ are often expensive in terms of time and compute %
   (e.g., training a deep neural network on a large dataset), so it is important to identify a good hyperparameter configuration $\xb_*$ with the least number of queries of $f$. Third, for complex models, the HP configuration space $\Xcal$ can have diverse types of attributes, some of which may be integer or categorical. For numerical attributes, search ranges need to be determined. Some attributes in $\Xcal$ can even be conditional (e.g., the width of the $l$-th layer of a neural network is only relevant if the model has at least $l$ layers). Finally, even if $f(\xb)$ varies smoothly in $\xb$, evaluations of $f(\xb)$ are typically noisy.

We present Amazon SageMaker Automatic Model Tuning (AMT), a fully managed system for gradient-free function optimization at scale.\footnote{\texttt{Amazon SageMaker} is a service that allows easy training and hosting of machine learning models. For details, see \url{https://aws.amazon.com/sagemaker} and~\citep{liberty2020elastic}.} The key contributions of our work are as follows:
\begin{itemize}
\item Design, architecture and implementation of hyperparameter optimization as a distributed, fault-tolerant, scalable, secure and fully managed service, integrated with Amazon SageMaker (\S\ref{sec:system}).
\item Description of the Bayesian Optimization algorithm powering AMT, including efficient hyperparameter representation, surrogate Gaussian process model, acquisition functions, and parallel and asynchronous evaluations (\S\ref{sec:algorithm}).
\item Overview of advanced features such as log scaling, automated early stopping and warm start (\S\ref{sec:advanced}).
\item Discussion of deployment results as well as challenges encountered and lessons learned (\S\ref{sec:results}).
\end{itemize}

\section{Preliminaries}\label{sec:prelim}
Traditionally, HPs are, either hand-tuned by experts in what amounts to a laborious process, or they are selected using brute force schemes such as grid search or random search. In response to increasing model complexity, a range of more sample-efficient HP optimization techniques have emerged. A comprehensive review of modern hyperparameter optimization is provided in \cite{Feurer:19}. Here, we will focus on work relevant in the context of AMT.

\subsection{Model-Free Hyperparameter Optimization}

Any HPO method is proposing evaluation points $\xb_1, \xb_2, \dots, \xb_T$, such that $$\min_{t=1,\dots, T} f(\xb_t) \approx \min_{\xb\in\Xcal} f(\xb).$$ For the simplest methods, the choice of $\xb_t$ does not depend on earlier observations. %
In {\em grid search}, we fix $K$ values for every HP $x_j$, then evaluate $f$ on the Cartesian product, so that $T = K^d$. In {\em random search}, we draw  $\xb_t$ independently and uniformly at random. More specifically, each $x_{t, j}$ is drawn uniformly from $\Xcal_j$. For numerical HPs, the distribution may also be uniform in a transformed domain (e.g., log domain). Random search is more effective than grid search when some of the HPs $x_j$ are irrelevant as it considers a larger number of values of the relevant ones \cite{Bergstra2012}. Both methods are easily parallelizable. Random search should always be considered as a baseline, and is frequently used to initialize more sophisticated HPO methods. Another alternative is picking a set of pseudo-random points known as Sobol sequences~\citep{Sobol1967}. The advantage is that they provide a better coverage of the search space, but are deterministic

These simple baselines can be improved upon by making use of earlier observations $\{ (\xb_{t_1}, f(\xb_{t_1}))\, |\,  t_1 < t \}$ in order to plan subsequent evaluations $\xb_{t_2}$, $t_2 \ge t$. Population-based methods, such as evolutionary or genetic algorithms, are attractive if parallel computing resources are available \cite{Hansen:01, Real:20}. In each generation, all configurations from a fixed-size population are evaluated. Then, a new population is created by randomly mutating a certain percentile of the top performing configurations, as well as sampling from a background distribution, and the process repeats with the next generation. Evolutionary algorithms (EAs) are powerful and flexible search strategies, which can work even in search spaces of complex structure. However, EAs can be difficult to configure to the problem at hand, and since fine-grained knowledge about $f$ can be encoded only in a large population, they tend to require a substantial amount of parallel compute resources. In some EAs, low-fidelity approximations of $f$ are employed during early generations \cite{Jaderberg2017} in order to reduce computational cost. Multi-fidelity strategies are discussed in more generality below.

\subsection{Surrogate Models. Bayesian Optimization}

A key idea to improve data efficiency of sampling is to maintain a {\em surrogate model} of the function $f$. At decision step $t$, this model is fit to previous data $\mathcal{D}_{<t} = \{ (\xb_{t_1}, f(\xb_{t_1}))\, |\, t_1 < t \}$. In {\em Bayesian optimization (BO)}, we employ probabilistic models, not only predicting best estimates (posterior means), but also uncertainties (posterior variances) for each $\xb\in\Xcal$: the value of the next $\xb_t$ could come from exploration (sampling where $f$ is most uncertain) or from exploitation (minimizing our best estimate), and a calibrated probabilistic model can be used to resolve the trade-off between these desiderata optimally \cite{Srinivas:12}. More concretely, we choose $\xb_t$ as the best point according to an acquisition function $\mathcal{A}(\xb | \mathcal{D}_{<t})$, which is a utility function averaged over the posterior predictive distribution $p(f(\xb) | \mathcal{D}_{<t})$. The most common surrogate model for BO is based on {\em Gaussian processes (GPs)} \cite{Rasmussen2006}, which not only have simple closed-form expressions for posterior and predictive distributions, but also come with strong theoretical guarantees. Other BO surrogate models include random forests \cite{Hutter2011} and Bayesian neural networks \cite{Springenberg2016}, which can suffer from uncalibrated uncertainty estimates. We give a detailed account of sequential BO with a GP surrogate model in Section~\ref{sec:algorithm}. Tutorials on BO are provided in \cite{Brochu2010, Shahriari2016}.

\subsection{Early Stopping. Multi-Fidelity Optimization}

Modern deep learning architectures can come with hundreds of millions of parameters, and even a single training run can take many hours or even days. In such settings, it is common practice to cheaply probe configurations $\xb$ by training for few epochs or on a subset of the data. While this gives rise to a low-fidelity approximation of $f(\xb)$, such data can be sufficient to filter out poor configurations early on, so that full training is run only for the most promising ones. To this end, we consider functions $f(\xb, r)$, $r$ being a resource attribute, where $f(\xb, r_{\text{max}}) = f(\xb)$ is the expensive metric of interest, while $f(\xb, r)$, $r < r_{\text{max}}$ are cheaper-to-evaluate approximations. Here, $r$ could be the number of training epochs for a deep neural network (DNN), the number of trees for gradient boosting, or the dataset subsampling ratio. When training DNNs, we can evaluate $f(\xb, r)$, $r=1,2,\dots$ by computing the validation metric after every epoch. For gradient boosting implementations supporting incremental training, we can obtain intermediate values $f(\xb, r)$ as well. In {\em early stopping} HPO, the evaluation of $\xb$ is terminated at a level $r$ if the probability of it scoring worse at $r_{\text{max}}$ than some earlier $\xb'$ is predicted high enough. The median rule is a simple instance of this idea \cite{Golovin2017}, while other techniques aim to extrapolate learning curves beyond the current $r$ \cite{Domhan2014, Klein2017}. Early stopping is particularly well suited to asynchronous parallel execution: whenever a job is stopped, an evaluation can be started for the next configuration $\xb$ proposed by HPO. An alternative to stopping configurations is to pause and (potentially) resume them later \cite{Swersky2014}. 

Besides early stopping HPO, successive halving (SH) \cite{Karnin:13, Jamieson2015} and Hyperband \cite{Li2016} are two other basic instances of {\em multi-fidelity} techniques.  In each round, $f(\xb, r_{\text{min}})$ is evaluated for a number $n$ of configurations $\xb$ sampled at random. Next, $f(\xb, 2 r_{\text{min}})$ is run for the top $n/2$ of configurations, while the bottom half are discarded. This filtering step is repeated until $r_{\text{max}}$ is reached. One drawback of SH and Hyperband is their synchronous nature, which is remedied by ASHA~\cite{Li:19}. All these methods make use of SH scheduling of evaluations, yet new configurations are chosen at random. BOHB \cite{Falkner2018} combines synchronous Hyperband with model-based HPO, and a more recent combination of ASHA with Bayesian optimization is MOBSTER \cite{abohb}. This method tends to be far more efficient than synchronous counterparts, and can often save up to half of resources compared to ASHA.

\section{The System}\label{sec:system}
In this section we give an overview of the system underpinning SageMaker AMT. We lay out design principles, describe the system architecture, and highlight a number of challenges related to running HPO in the cloud.

\subsection{Design Principles}\label{sec:principles}

We present the key requirements underlying the design of SageMaker AMT.
\begin{itemize}
\item {\em Easy to use and fully managed}: To ensure broad adoption by data scientists and ML developers with varying levels of ML knowledge, AMT should be easy to use, with minimal effort needed for a new user to setup and execute. Further, we would like AMT to be offered as a fully managed service, with stable API and default configuration settings so that the implementation complexity is abstracted away from the customer. AMT spares users the pain to provision hardware, install the right software, and download the data. It takes care of uploading the models to the users’ accounts and providing them with training performance metrics. 
\item {\em Tightly integrated with other SageMaker components}: Considering that model tuning (HPO) is typically performed as a part of the ML pipeline involving several other components, AMT should seamlessly operate with other SageMaker components and APIs.
\item {\em Scalable}: AMT should scale with respect to different data sizes, ML training algorithms, number of HPs, metrics, HPO methods and hardware configurations. Scalability includes a failure-resistant workflow with built-in retry mechanisms to guarantee robustness.
\item {\em Cost-effective}: AMT should be cost-effective to the customer, in terms of both compute and human costs. We would like to enable the customer to specify a budget and support cost reduction techniques such as early stopping and warm start.
\item {\em Available}: A pillar of AMT's design is to ensure that it is highly available. This includes success of the synchronous API's as well as minimizing the failures of the asynchronous workflows underlying the hyperparameter tuning jobs.
\item {\em Secure}: Security is a top priority for AMT. In the context of HPO, this means not putting customer's data at risk. As we will elaborate below, our system does not store any customer data. In addition, we provide extra security features such as the ability to run workloads in a network-isolated mode (i.e., containers cannot make network calls) and/or in a VPC, addressing security requirements for AMT's users.
\end{itemize}

\subsection{System Architecture}\label{sec:arch}
Sagemaker AMT provides a distributed, fault-tolerant, scalable, secure and fully-managed service for HPO. One of the key building blocks of the service is the AWS Sagemaker Training platform, which executes the training jobs and obtains the value of the objective metric for any candidate hyperparameter chosen by the Hyperparameter Selection Service. In this way, each candidate set of hyperparameters tried is associated to a corresponding Sagemaker Training Job in the user’s AWS account. Using the SageMaker training platform allows AMT to scale well. Several training jobs can be run in parallel, use distributed clusters of EC2 instances, and scale HPO to large volumes of data. This includes a failure-resistant workflow with built-in retry mechanisms to guarantee robustness.

\begin{figure}[t]
\center
   \includegraphics[width = 0.5\textwidth]{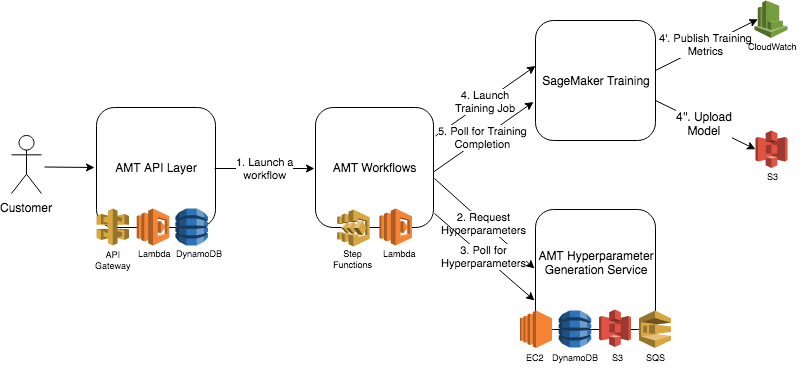}
   \vspace{-0.3cm}
\caption{System architecture of SageMaker AMT. The diagram shows the different components of AMT and how they interact. A number of AWS building blocks are combined to deliver a scalable, robust, secure and fully managed service.}
\vspace{-0.3cm}
\label{fig:architecture}
\end{figure}

Sagemaker  AMT's  backend  is  built  using  a  fully  server-less architecture by means of a number of AWS building blocks. It uses AWS  API  Gateway,  AWS  Lambda,  AWS  DynamoDB, AWS Step Functions, AWS Cloudwatch Events in  its back-end  workflow.   AWS  API  Gateway  and AWS Lambda is used to power the API Layer which customers use to call various Sagemaker AMT APIs, such as Create/List/Describe/StopHyperparamaterTuningJobs.  AWS DynamoDB is used as the persistent store to keep all the metadata associated with the job and also track the current state of the job. The overall system's architecture is depicted in Figure~\ref{fig:architecture}. Sagemaker AMT deals only with the metadata for the jobs and all customer data is handled by the Sagemaker Training platform, ensuring that no customer data is stored into the DynamoDB Table. AWS Cloudwatch Events, AWS Step Functions and AWS Lambda are used in the AMT workflows engine, which is responsible for kicking off the evaluation of hyperparameter configurations from the Hyperparameter Selection Service, starting training jobs, tracking their progress and repeating the process until the stopping criterion is met.

\subsection{Challenges for HPO in the cloud}
While running at scale poses a number of challenges, AMT is a highly distributed and fault tolerant system. System resiliency was one of the guiding principles when building AMT. Example failure scenarios are when the BO engine suggests hyperparameters that can run out of memory or when individual training jobs fail due to dependency issues. The AMT workflow engine is designed to be resilient against failures, and has a built-in retry mechanism to guarantee robustness.%

AMT runs every evaluation as a separate training job on the SageMaker Training platform. Each training job provides customers with a usable model, logs and metrics persisted in CloudWatch. A training job involves setting up a new cluster of EC2 instances, waiting for the setup to complete, and downloading algorithm images. This introduced an overhead that was pronounced for smaller datasets. To address this, AMT puts in place compute provisioning optimizations to reduce the time of setting up clusters and getting them ready to run the training.

\section{The Algorithm}\label{sec:algorithm}
Next, we describe the main components and modelling choices behind the BO algorithm powering SageMaker AMT. We start with the representation of the %
hyperparameters, followed by the key decisions we made around the choice of the surrogate model and acquisition function. We also describe the way AMT tackles the common use case of exploiting parallel compute resources.

\subsection{Input configuration}
To tune the HPs of a machine learning model, AMT needs an input description of the space over which the optimization is performed. We denote by $\Hcal$ the set of HPs we are working with, such as $\Hcal = \{ \texttt{learning rate}, \texttt{loss function}\}$.
For each $h \in \Hcal$, we also define  $\Bcal_h$ the domain of values that $h$ can possibly take, leading to the global domain of HPs
$
\Bcal_{\Hcal}  \defin \Bcal_{h_1} \times \dots \times \Bcal_{h_{|\Hcal|}}.
$
Each $h\in\Hcal$ has data type continuous (real-valued), integer, or categorical, where the numerical types come with lower and upper bounds. Integer HPs are handled by working in the continuous space and rounding to the nearest integer, while categorical HPs are one-hot encoded.

It is common among ML practitioners to exploit some prior knowledge to adequately transform the space of some HP, such as applying a log-transform for a regularisation parameter. This point is important as the final performance hinges on this preprocessing. There has been a large body work dedicated to automatically finding such input transformations (e.g., see \cite{Assael2014} and references therein). In AMT this is tackled through input warping and log scaling, which are described in the following sections.

\subsection{Gaussian process modelling}
Once the input HPs are encoded, AMT builds a model mapping hyperparameter configurations to their predicted performance. We follow a form of GP-based global optimisation~\citep{Jones1998,Lizotte2008,Osborne2009,Brochu2010}, meaning that the %
function $f$ to minimize is assumed to be drawn from a GP with a given mean and covariance function. The GP is a particularly suitable model for an online learning system such as AMT, where past hyperparameter evaluations are iteratively used to pick the next ones. Since observations $y$ collected from $f$ are normalized to mean zero, we can consider a zero-mean function without loss of generality. The choice of the covariance function $\Kcal_\thetab$, which depends on some Gaussian process hyperparameters (GPHPs) $\thetab$, will be discussed in detail in the next paragraph.

More formally, and given an encoded configuration $\xb$, our probabilistic model reads $f(\xb)  \sim \Gcal\Pcal(0,\Kcal_\thetab)$ and $y | f(\xb)  \sim  \Ncal(f(\xb),\sigma_0^2)$, where the observation $y$ is modelled as a Gaussian random variable with mean $f(\xb)$ and variance $\sigma_0^2$: a standard GP regression setup~\citep[Chapter 2]{Rasmussen2006}. Many choices for the covariance function (or kernel) $\Kcal_\thetab: \Real^d \times \Real^d \mapsto \Real$ are possible. The Mat\'ern-$5/2$ kernel with automatic relevance determination (ARD) parametrisation~\citep[Chapter 4]{Rasmussen2006} is advocated in~\citep{Snoek2012,Swersky2013,Snoek2014}, where it is shown that ARD does not lead to overfitting, provided that the GPHP are properly handled. We follow this choice, which has become a de-facto standard in most BO packages.

\paragraph{GP hyperparameters}
Our probabilistic model comes with some GPHPs $\thetab$. We highlight two possible options to treat these parameters, both of which are implemented in AMT. A traditional way of determining the GPHPs consists in finding $\thetab$ that maximises the log marginal likelihood of our probabilistic model~\citep[Section 5.4]{Rasmussen2006}. While this approach, known as empirical Bayes, is efficient and often leads to good results, \cite[and follow-up work]{Snoek2012} rather advocate the full Bayesian treatment of integrating out $\thetab$ by Markov chain Monte Carlo (MCMC). In our experiments we found the latter approach to be less likely to overfit in the few-observation regime (i.e., early in the BO procedure), but is also more costly, since GP computations have to be done for every MCMC sample. In AMT, we implement slice sampling, one of the most widely used techniques for GPHPs~\citep{Murray2010, Mackay2003}. In our implementation we use one chain of 300 samples, with 250 samples as burn-in and thinning every 5 samples, resulting in an effective sample size of 10. We fix upper and lower bounds on the GPHPs for numerical stability, and use a random (normalised) direction, as opposed to a coordinate-wise strategy, to go from our multivariate problem ($\thetab\in \Real^k$) to the standard univariate formulation of slice sampling. We observed that slice sampling is a better approach to learn the GPHPs compared to empirical Bayes, especially at the beginning of HPO where the latter is more prone to overfitting due to the small number of observations.

\paragraph{Input warping}
It is common practice among ML practitioners to exploit some prior knowledge to adequately transform the space of some HP, such as applying a log-transform for a regularization parameter. We leverage the ideas developed in \cite{Snoek2014}, where the configuration $\xb$ is transformed entry-wise by applying, for each dimension $j \in \{1, \dots, d\}$, $\omega(\xb_j) \defin \text{BetaCDF}(\xb_j,\alpha_j,\beta_j)$,
where $\{\alpha_j,\beta_j\}_{j\in\{1, \dots, d\}}$ are GPHPs that govern the shape of the transformation. We refer to $\omegab(\xb) \in \Real^d$ as the vector resulting from all entry-wise transformations. An alternative, which is the default choice in AMT, is to consider the CDF of the Kumaraswamy's distribution, which is more tractable than the CDF of the Beta distribution.
A convenient way of handling these additional GHPHs is to overload our definition of the covariance function so that for, any two $\xb,\xb'$, $K_\thetab(\xb,\xb) \defin K_\thetab(\omegab(\xb),\omegab(\xb'))$, where $\{\alpha_j,\beta_j\}_{j\in \{1, \dots, d\}}$ are merged within the global vector $\thetab$ of GPHPs. %

\subsection{Acquisition functions}\label{sec:acquisitionfunction}
Denote evaluations done up to now by $\Dcal = \{  (\xb^c, y^c)\; |\; c\in\Ccal \}$. In GP-based Bayesian optimization, an \textit{acquisition function} is optimized in order to determine the hyperparameter configuration at which $f$ should be evaluated next. Most common acquisition functions $\Acal(\xb)$ depend on the GP posterior $P(f\; |\; \Dcal)$ only via its marginal mean $\mu(\xb)$ and variance $\sigma^2(\xb)$. There is an extensive literature dedicated to the design of acquisition functions. The Expected improvement (EI) was introduced by \cite{Mockus1978} and is probably the most popular acquisition function (notably popularised by the EGO algorithm \cite{Jones1998}). EI is the default choice for toolboxes like SMAC \citep{Hutter2011}, Spearmint \citep{Snoek2012}, and AMT. It is defined as $\Acal(\xb | \Dcal, \thetab) = \Exp[  \max\{0,y^* - y(\xb) \} ]$, where the expectation is taken with respect to the posterior distribution $y(\xb) \sim \Ncal(\mu(\xb), \sigma^2(\xb))$. Here, $y^* = \min\{ y^c\; |\; c\in \Ccal \}$ is the best target value observed so far. In the case of EI, the expectation appearing in $\Acal(\xb)$ is taken with respect to a Gaussian marginal consistent with the Gaussian process posterior and, as such, has a simple closed-form expression. Another basic acquisition function is provided by Thompson sampling, which is optimized by drawing a realization of the GP posterior and searching for its minimum point $\xb$~\citep{Thompson1933, Hoffman2014}. Exact Thompson sampling requires a sample from the {\em joint} posterior process, which is intractable. An approximation to this scheme is used in AMT, whereby {\em marginal} variables are sampled from $\Ncal(\mu(\xb), \sigma^2(\xb))$ at $\xb$ locations from a dense set. The set is obtained through a Sobol sequence generator~\citep{Sobol1967} populating the search space as densely as possible. Rather than using the sampled variables directly, the resulting pseudo-random grid is used as a set of anchor points to initialize the local optimization of the EI. This scales linearly in the number of locations and works well in practice. 

Other interesting families of acquisition functions include those built on upper-confidence bound ideas \citep{Srinivas2009} or information-theoretic criteria \citep{Villemonteix2009,Hennig2012,Hernandez-Lobato2014,Wang2017}. While more sample-efficient, many of these typically come with higher computational cost than the EI, which adds overhead when the tuned model is fast to train (e.g., on a small dataset). For this reason, we chose the EI as a robust acquisition function striking a balance between performance, simplicity and computational time. Alternative acquisition functions to make the EI cost-aware and steer the hyperparameter search towards cheaper configurations are described in \citep{Lee2020, guinet2020}. These variants mitigate the cost associated to expensive hyperparameter configurations, such as very large architectures when tuning neural network models.

\subsection{Parallelism}

While the BO procedure presented so far is purely sequential, many users are interested in exploiting parallel resources to speed up their tuning jobs. To address this, AMT lets them specify whether the %
target function should be evaluated in parallel for different candidates, and with how many parallel evaluations at a time. There are different ways of taking advantage of a distributed computational environment. If we suppose we have access to $L$ threads/machines, we can simply return the top-$L$ candidates as ranked by the acquisition function $\Acal$. We proceed to the next step only when the $L$ candidate's evaluations in parallel are completed. This strategy works well as long as the candidates are diverse enough and their evaluation time is not too heterogenous, which is unfortunately rarely the case. For these reasons, AMT optimizes the acquisition function so as to induce diversity and adopts an asynchronous strategy. As soon as one of the $L$ evaluations is done, we update the GP with this new configuration and pick the next candidate to fill in the available computation slot (making sure, of course, not to select one of the $L-1$ pending candidates). One disadvantage is that this does not take into account the information coming from the fact that we picked the $L-1$ pending candidates. To tackle this, asynchronous processing could %
be based on fantasizing~\cite{Snoek2012, abohb}.

\section{Advanced features} \label{sec:advanced}
Beyond standard BO, AMT comes with a number of extra features to speed up tuning jobs, saving computational time and potentially finding better hyperparameter configurations. We start by describing log scaling, followed by early stopping and warm-starting.

\subsection{Log Scaling} 
\label{sec:log-scale}

\begin{figure}[t]
   \includegraphics[width = 0.35\textwidth]{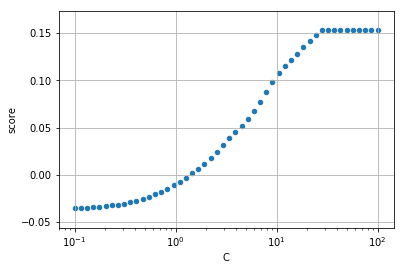}
   \vspace{-0.5cm}
\caption{Change of validation set score with different values of the capacity parameter of a Support Vector Machine. Note the logarithmic scale of the capacity parameter $C$.}
\vspace{-0.3cm}
\label{fig:logscale}
\end{figure}

A common property of learning problems is that a linear change in validation performance requires an exponential increase in the learning capacity of the estimator (as defined by the VC dimension in \cite{vapnik2013nature}). To illustrate this, Figure \ref{fig:logscale} shows the relationship between the capacity parameter of SVM and validation accuracy. As a result, a wide search range is often chosen for hyperparameters controlling model capacity. For instance, a typical choice for the capacity parameter $C$ of support vector machine is $\{ 10^{-9} \dots 10^{9} \}$. Note that 99\% of the volume of this example search space corresponds to values of hyperparameter $C \in \{10^{7} \dots 10^{9} \}$. As a result, smaller values of $C$ might be under-explored by BO if applied directly to such range, as it attempts to evenly explore the search space. To avoid under-exploring smaller values, a log transformation is applied to such model capacity related variables. Such transformation is generally referred to as ``log scaling''. Search within transformed search spaces can be done automatically by AMT, provided that the user indicates that such transformation is appropriate for a given hyperparameter via the API. For all the algorithms provided in SageMaker, such as XGBoost and Linear Learner, recommendations are given for which hyperparameters log scaling is appropriate and accelerates the tuning. 

Compared to the GP input warping, which automatically detects the right scaling as the tuning job progresses, log scaling is applied from the start and comes without extra parameters to estimate. The internal warping can learn a larger family of transformations compared to the ones supported by hyperparameter scaling, and is thus always active by default. However, log scaling is useful when the appropriate input transformation is known in advance. Unlike input warping, it can be used not only with BO but also with random search.

\subsection{Early Stopping}
Tuning involves evaluating several hyperparameter configurations, which can be costly. Generally, a new hyperparameter configuration $\xb_t$ may not improve over the previous ones, i.e., $f(\xb_{t}) \geq f(\xb_{t'})$ for one or more $t' < t$. With early stopping, AMT uses the information of the previously-evaluated configurations to predict whether a specific candidate $\xb_t$ is promising. If not, it stops the evaluation, thus reducing the overall time. This works whenever we can obtain intermediate values $f(\xb_t^1), f(\xb_t^2), ..., f(\xb_t^n)$, with $f(\xb_t^r)$ representing the value of the objective for the configuration $\xb_t$ at training iteration $r$. 

To implement early stopping, AMT employs the simple but effective median rule~\citep{Golovin2017} to determine which HP configurations to stop early. If $f(\xb_t^r)$ is worse than the median of the previously evaluated configurations at the same iteration $r$, we stop the training. An alternative is to predict future performance via a model and stop poor configurations. In our experiments, the simple median rule performed at least as well as, and often better, than predictions from linear and random forest models. A concern with the median rule is that lower-fidelities are not necessarily representative of the final values: as the training proceeds, a seemingly poor HP configuration can eventually improve enough to become the best one. To improve resilience, we only make stopping decisions after a given number of training iterations. As the total number of iterations can vary for different algorithms and use-cases, this threshold is determined dynamically based on the duration of the fully completed hyperparameter evaluations. We also considered the additional safeguard of always completing 10 hyperparameter evaluations before activating the median rule. However, this produced only slight improvements in the final objective value at the price of reduced time savings, and was therefore discarded.

\subsection{Warm start}
A typical ask from users running several related tuning jobs is to be able to build on previous experiments. For example, one may want to gradually increase the number of iterations, change hyperparameter ranges, change which hyperparameters to tune, or even tune the same model on a new dataset. In all these cases, it is desirable to re-use information from previous tuning jobs rather than starting from scratch. Since related tuning tasks are intuitively expected to be similar, this setting lends itself to some form of \textit{transfer learning}. With warm start, AMT uses the results of previous tuning jobs to inform which combinations of hyperparameters to search over in the new tuning job. This can make the search for the best combination of hyperparameters more efficient.

Speeding up HP tuning with transfer learning is an active line of research. Most of this activity has focused on transferring HP knowledge across different datasets~\cite{Bardenet2013,Feurer2015,Golovin2017,perrone_multiple_2017,Perrone2018,Perrone2019,Salinas2020}. However, most of this previous work assumes the availability of some meta-data, describing in which sense datasets differ from each other~\cite{Feurer2015}. When designing warm start, we found this assumption prohibitively restrictive in practice. Computing meta-data in real-world predictive systems is challenging due to the computational overhead or privacy reasons. We thus opted for a light-weight solution, purely based on past hyperparameter evaluations and requiring no access to meta-data. It should be noted that simple warm-starting typically works well when data is stationary, and can otherwise bias the surrogate model towards unpromising configurations. Future improvements could make the solution more robust to ``negative transfer'' arising from tasks that are not positively correlated.

\subsection{AMT for AutoML}
AMT also serves as key component for SageMaker Autopilot~\cite{PDas2020}, the AutoML service of Amazon SageMaker. In Autopilot, users can upload a tabular dataset and the service will explore a complex search space, consisting of feature preprocessing, different ML algorithms and their hyperparameter spaces. For more details on how AMT is used in Autopilot we refer the readers to~\cite{PDas2020}. The application to Autopilot highlights how finding a good single ML model through hyperparameter optimization is still key for real-world applications, including in the context of AutoML. Many techniques, such as ensembling~\cite{caruana2004ensemble}, can boost the accuracy by combining different models. However, when the number of models in an ensemble is large, inference time may become unacceptable, hindering practical usage. The application of AMT to Autopilot demonstrates the relevance of HPO in real-world applications, which require simple, robust and deployable models. Further, HPO is not incompatible with ensembling, and can be applied to build better ensembles.

\section{Use Cases and Deployment Results}\label{sec:results}
We now consider a number of case studies highlighting the benefits of AMT, with respect to both its core algorithm and advanced functionalities. We first demonstrate the appeal of the BO strategy implemented in AMT over random search, and then turn to the empirical advantages of the advanced features we described in the previous section.

\subsection{BO vs random search}

\begin{figure*}[t]
        \includegraphics[width = 0.50\textwidth]{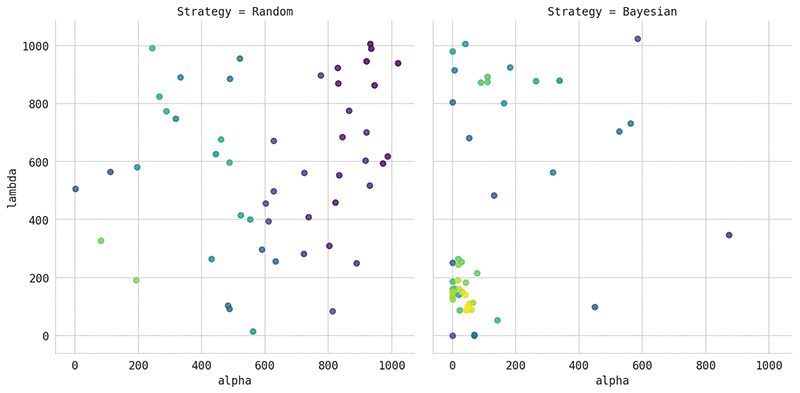}
        \includegraphics[width = 0.40\textwidth]{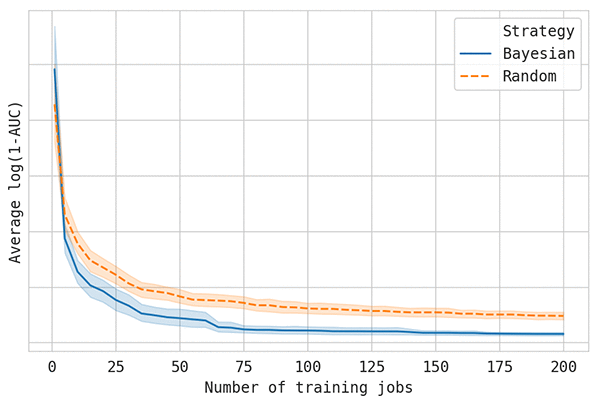}
         \vspace{-0.3cm}
    \caption{\texttt{Left}: Hyperparameters suggested by random search. The $x$-axis and $y$-axis are the hyperparameter values for \texttt{alpha} and \texttt{lambda} and the color of the points represent the quality of the hyperparameters (yellow: better AUC; violet: worse AUC).  
    \texttt{Middle}: Hyperparameters are suggested by AMT through BO. 
    \texttt{Right}: Best model score obtained so far ($y$-axis, where lower is better) as more hyperparameter evaluations are performed ($x$-axis).}
     \vspace{-0.3cm}
    \label{fig:rs-bo}
\end{figure*}

A very common use case is tuning regularization hyperparameters. We demonstrate this by tuning the regularization terms \texttt{alpha} and \texttt{lambda} of the XGBoost algorithm on the \texttt{direct marketing} binary classification dataset from UCI to minimize the AUC. Figure \ref{fig:rs-bo} compares the performance of random search and BO, as implemented in AMT. Each experiment was replicated with 50 different random seeds, and we reported average and standard deviation. Specifically, from the left and middle plots in Figure \ref{fig:rs-bo}, we can see that BO suggested better performing hyperparameters then random search. The right plot of Figure \ref{fig:rs-bo} shows that BO consistently outperforms random search across all number of hyperparameter evaluations.\footnote{A notebook to run this example on AWS SageMaker: \url{https://github.com/awslabs/amazon-sagemaker-examples/tree/master/hyperparameter_tuning/xgboost_random_log}.} Although BO is more competitive in the sequential setting, AMT also offers random search. It is important to note that users find random search particularly useful in the distributed setting where a large number of configurations are evaluated in parallel, reducing wall-clock time.

\subsection{Log scaling}
Figure \ref{fig:rs-bo} depicts a run of AMT with log scaling. While BO focuses on a small region of the search space, it still attempts to explore the rest of the search space. This suggests that well performing training jobs are found with low values of \texttt{alpha}. With log scaling, BO is steered towards these values and focuses on the most relevant region of the space. Models with larger learning capacity require more compute resources to perform the training phase (e.g., larger depth of trees in XGBoost leads to exponentially larger training time). Log scaling accelerates the search of good hyperparameter configurations, and also reduces the exploration of costly configurations -- which usually correspond to large hyperparameter values. As the choice of scaling requires in-depth knowledge of the internals of the learning algorithms, recommendations for common algorithms on SageMaker were made available. A lesson learned from the development of log scaling is to carefully consider all possible edge-case inputs to the service. We uncovered a non-trivial failure of AMT where a user would first run a tuning job with a linear scaling of a hyperparameter (e.g., in $[0.0, 1.0]$), and use it to warm start a new tuning job with log scaling enabled. While the value $0$ could have been explored in the parent job, this becomes invalid in the child job. These non-trivial issues can only be uncovered through careful examination of all possible edge-cases associated to a new functionality.

\subsection{Early stopping}
\begin{figure}[t]
        \includegraphics[width = 0.23\textwidth, trim={0 0 0.5cm 0},clip]{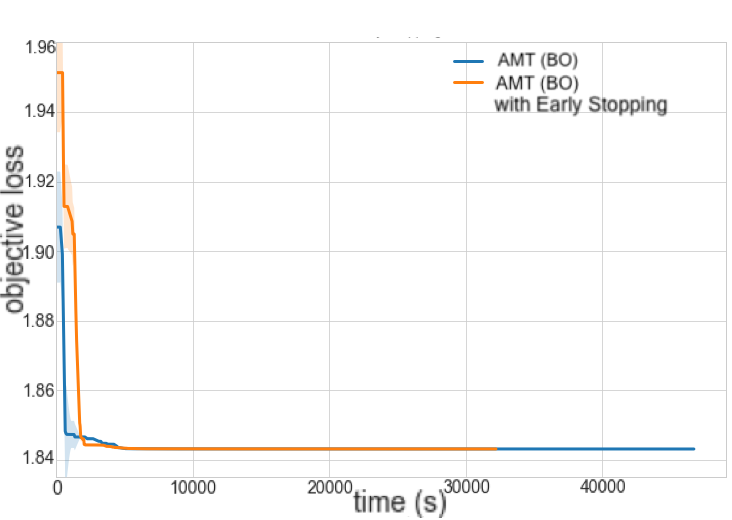}
    \includegraphics[width = 0.23\textwidth, trim={0 0 0.5cm 0},clip]{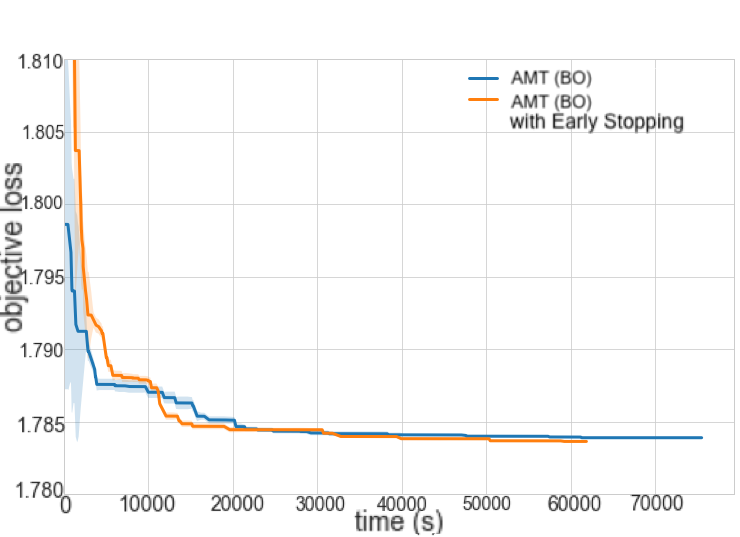}
        \vspace{-0.2cm}
    \caption{Absolute loss over time with and without early stopping when tuning linear learner on the Gdelt data, for single instance mode (left) and distributed mode (right). In both cases, early stopping achieves a similar loss in less time.}
    \vspace{-0.5cm}
    \label{fig:early-stopping}
\end{figure}

One of the main concerns with early stopping is that it could negatively impact the final objective values. Furthermore, the effect of early stopping is most noticeable on longer training jobs. We consider Amazon SageMaker's built-in linear learner algorithm on the Gdelt (Global Database of Events, Language and Tone) dataset in both single instance and distributed training mode.\footnote{\url{http://www.gdeltproject.org/}.}  Specifically, we used the Gdelt dataset from multiple years in distributed mode and from a single year in single instance mode. Figure \ref{fig:early-stopping} compares a hyperparameter tuning job with and without early stopping. Each experiment was replicated 10 times, and the median of the best model score so far (in terms of absolute loss, lower is better) is shown on the $y$-axis, while the $x$-axis represents time. Each tuning job was launched with a budget of 100 hyperparameter configurations to explore. AMT with early stopping not only explores the same number of HP configurations in less time, but yields hyperparameter configurations with similar performance.

\subsection{Warm start}
\begin{figure}
    \includegraphics[width = 0.35\textwidth, trim={0 0 0.5cm 0},clip]{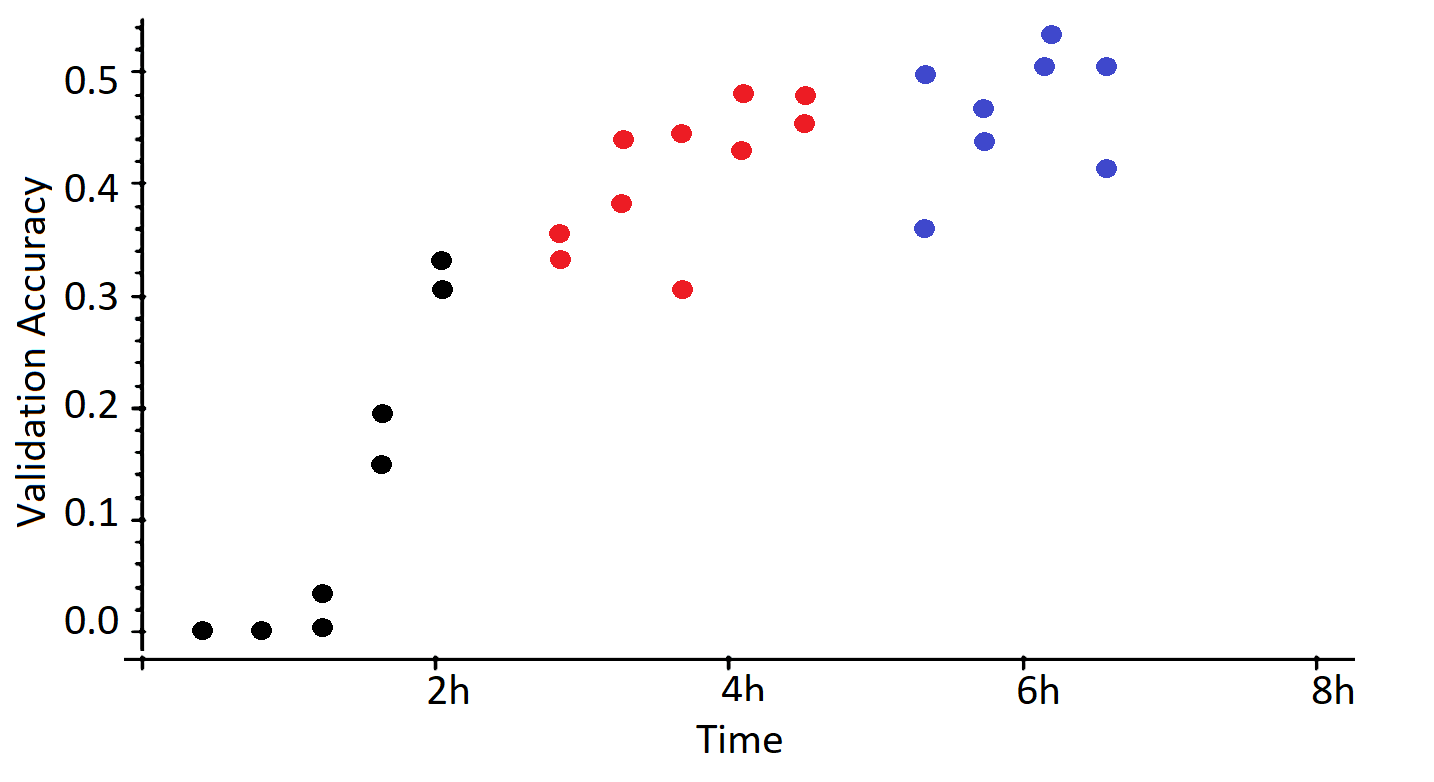}
        \vspace{-0.45cm}
    \caption{Validation accuracy ($y$-axis) over time ($x$-axis) for the tuned image classifier on the Caltech-256 dataset. The black dots show some iterations of AMT when starting from scratch. Warm start allows you to keep tuning the same algorithm on the same data (red dots) and on a transformed version of the data obtained via data augmentations (blue dots), with the validation accuracy continuously improving.}
     \vspace{-0.3cm}
    \label{fig:warm-start}
\end{figure}

It is common to update models regularly, typically with a different choice of the hyperparameter space, iteration count or dataset. This requires re-tuning the model hyperparameters. AMT's warm start offers a simple approach to learn from previous tuning tasks. We demonstrate this on the problem of building an image classifier and iteratively tuning it by running multiple hyperparameter tuning jobs. We focus on two simple use cases: running two sequential hyperparameter tuning jobs on the same algorithm and dataset, and launching a new tuning job on the same algorithm on an augmented dataset. We train \texttt{Amazon SageMaker}'s built-in image classification algorithm on the Caltech-256 dataset, and tune its hyperparameters to maximize validation accuracy. Figure~\ref{fig:warm-start} shows the impact of warm starting from previous tuning jobs. Initially, there is a single tuning job. Once complete, we launch a new tuning job by selecting the previous job as the parent. The plot shows that the new tuning job (red dots) quickly detects good hyperparameter configurations thanks to the knowledge from the parent job (black dots). As the optimization progresses, the validation accuracy reaches 0.47, thus improving over 0.33, the best previous metric found by running the tuning job from scratch. Warm start can also be applied \emph{across different datasets}. We apply a range of data augmentations, including crop, color, and random transformations (i.e., image rotation, shear, and aspect ratio variations). To launch the last tuning job, we warm start from both previous jobs and run BO for 10 more iterations. The accuracy for the new tuning job (blue dots) improved again compared to the parent jobs (red and black dots), reaching 0.52.

As expected, this feature particularly helped users who run their tuning jobs iteratively. Interestingly, we also found this feature to be a practical enabler for users who would like to tune their models for a very large number of evaluations. While this would be prohibitively expensive due to the cubical scaling of GPs, it can be achieved through running tuning jobs in a sequence, each time warm-starting from the previous one (e.g., with 500 HPO evaluations per tuning task).

\subsection{Post launch performance}
Many customers have adopted AMT as part of their ML workflows since the product launch in December 2017. In particular, four use cases by large enterprises for AMT are currently publicly presented as example use cases.\footnote{\url{https://aws.amazon.com/sagemaker/customers}} Applications include automated bidding to consume or supply electrical energy, click through rate and conversion rate predictions, malicious actor detection for DNS security, as well as general prototyping of novel ML applications. Note that all the listed applications involve business critical data for customers of AMT, which highlights the importance of building a tuning service on top of a secure training platform to guarantee secure storage and processing of data. The generality of our container-based tuning service allows us to accommodate a wide variety of learning algorithms, such as custom  neural-network-based solutions implemented in TensorFlow, or gradient boosting with decision trees.

An abundance of empirical data is continuously collected to monitor AMT's performance. For instance, API communication was available at more than service-level agreement for the 99.99\% of time in 2020. A number of requested tuning jobs also showcased the scalability of service. An example includes requests with 5 training jobs in parallel, each running on a cluster of 100 CPU machines of 4 cores and 16 GB of RAM. Other examples are individual training jobs involving clusters of 128 GPU accelerators. The overall service is able to process spikes of many hundreds of tuning jobs with no operational issues.

\section{Related Work} 

Before concluding, we briefly review open source solutions for %
gradient-free optimization in this section. Rather than providing an exhaustive list, we aim to give an overview of the tools available publicly. One of the earliest packages for Bayesian Optimization using GP as a surrogate model is \texttt{Spearmint} \cite{Snoek2012}, where several important extensions including multi-task BO \cite{Swersky2013}, input-warping \cite{Snoek2014} and handling of unknown constraints \cite{Gelbart2014} have been introduced. The same strategy has also been implemented in other open source packages such as \texttt{BayesianOptimization} \cite{Nogueira2014}, \texttt{scikit-optimize} (easy to use when training scikit-learn algorithms) and \texttt{Emukit} \cite{emukit2019}.\footnote{\url{https://github.com/scikit-optimize}}  Unlike using a GP as surrogate model, \texttt{SMAC} \cite{hutter2011sequential} uses random forest, which makes it appealing for high dimensional and discrete problems. With the growing popularity of deep learning frameworks, BO has also been implemented for all the major deep learning frameworks. \texttt{BoTorch} \cite{balandat2019botorch} is the BO implementation built on top of \texttt{PyTorch} \cite{paszke2017automatic} and \texttt{GPyTorch} \cite{gardner2018gpytorch}, with an emphasis on modular interface, support for scalable GPs as well as multi-objective BO. In \texttt{TensorFlow}, there is \texttt{GPflowOpt} \cite{GPflowOpt2017}, which is the BO implementation dependent on \texttt{GPflow}  \cite{GPflow2017}.\footnote{\url{https://www.tensorflow.org/}} Finally, \texttt{AutoGluon HNAS} \cite{abohb} provides asynchronous BO, with and without multi-fidelity optimization, as part of \texttt{AutoGluon}.%

\section{Conclusion}
We presented SageMaker AMT, a fully-managed service to optimize gradient-free functions in the cloud. 
Powered by BO, Sagemaker AMT brings together the state-of-the-art for hyperparameter optimization and offers it as a highly scalable solution. We gave insights into its design principles, showed how it  integrates with other SageMaker's components, and shared practice for designing a fault-tolerant system in the cloud. Through a set of real-world use cases, we showed that AMT is an effective tool for HPO. It also offers a set of advanced features, such as automatic early stopping and warm-starting from previous tuning jobs, which demonstrably speed up the search of good hyperparameter configurations. 
Beyond performance metrics, several ML applications involve optimizing multiple constraints and alternative metrics at the same time, such as maximum memory usage, inference latency or fairness~\citep{Perrone2019mes, Perrone2020, Lee2020, guinet2020}. In the future, AMT could be extended to optimize multiple objectives simultaneously, automatically suggesting hyperparameter configurations that are optimal along several criteria and search for the Pareto frontier of the multiple objectives. 

\begin{acks}
AMT has contributions from many members  of the SageMaker team, notably Ralf Herbrich, Leo Dirac, Michael Brueckner, Viktor Kubinec, Anne Milbert, Choucri Bechir, Enrico Sartoriello, Thibaut Lienart, Furkan Bozkurt, Ilyes Khamlichi, Yifan Su, Adnan Kukuljac, Ugur Adiguzel, Mahdi Heidari.
\end{acks}

\bibliographystyle{abbrv}
\bibliography{references,bibliography,MainBibliography}

\end{document}